\documentclass[10pt,twocolumn,letterpaper]{article}

\usepackage[pagenumbers]{wacv} %

\usepackage{graphicx}
\usepackage{amsmath}
\usepackage{amssymb}
\usepackage{booktabs}

\usepackage[accsupp]{axessibility} %

\usepackage[pagebackref,breaklinks,colorlinks]{hyperref}

\usepackage[capitalize]{cleveref}
\crefname{section}{Sec.}{Secs.}
\Crefname{section}{Section}{Sections}
\Crefname{table}{Table}{Tables}
\crefname{table}{Tab.}{Tabs.}

\def\wacvPaperID{790} %
\def\confName{WACV}
\def\confYear{2025}

\usepackage{amssymb}
\usepackage{amsfonts}
\usepackage{amsmath}
\usepackage{amsxtra}
\usepackage{cancel}
\usepackage{dsfont}
\usepackage{graphicx}
\usepackage{tikz}
\usepackage{tikz-qtree}
\usetikzlibrary{shapes}
\usetikzlibrary{positioning}
\usetikzlibrary{trees}
\usepackage{mathcomp}
\usepackage{mathtools}
\usepackage{multirow}
\usepackage{verbatim}
\usepackage{polynom}
\usepackage{textcomp}
\usepackage{float}
\usepackage{pdflscape}
\usepackage{csquotes}
\usepackage{afterpage}
\usepackage{makecell}
\usepackage{listings}
\usepackage{enumitem}
\usepackage{minibox}
\usepackage{algpseudocode}
\usepackage{shuffle}
\usepackage{svg}
\usepackage{pifont}
\usepackage{subcaption}
\usepackage{siunitx}
\usepackage{microtype}
\usepackage{placeins}

\DeclareMathSymbol{\mlq}{\mathord}{operators}{``}
\DeclareMathSymbol{\mrq}{\mathord}{operators}{`'}

\newcommand{\R}{\mathbb R}
\newcommand{\N}{\mathbb N}

\newcommand{\eps}{\varepsilon}

\newcommand{\id}{\mathrm{id}}

\renewcommand{\phi}{\varphi}

\newcommand{\softmax}{\operatorname{softmax}}

\renewcommand{\O}{\mathcal{O}}

\newcommand{\inlinecode}[1]{\texttt{#1}}

\begin{document}
\title{Which Transformer to Favor: \\ A Comparative Analysis of Efficiency in Vision Transformers}

\author{Tobias Christian Nauen$^{1,2}$\\
{\tt\small tobias\_christian.nauen@dfki.de}
\and
Sebastian Palacio$^3$\\
{\tt\small sebastian.palacio@de.abb.com}
\and
Federico Raue$^{2}$\\
{\tt\small federico.raue@dfki.de}
\and
Andreas Dengel$^{1,2}$\\
{\tt\small andreas.dengel@dfki.de}
\and
${}^1$University of Kaiserslautern-Landau, Gottlieb-Daimler-Straße, Kaiserslautern\\
${}^2$German Research Center for Artificial Intelligence (DFKI), Trippstadter Straße 122, Kaiserslautern\\
${}^3$ABB AG, Kallstadter Str. 1, 68309 Mannheim
}

\maketitle

\begin{abstract}
	Self-attention in Transformers comes with a high computational cost because of their quadratic computational complexity, but their effectiveness in addressing problems in language and vision has sparked extensive research aimed at enhancing their efficiency.
	However, diverse experimental conditions, spanning multiple input domains, prevent a fair comparison based solely on reported results, posing challenges for model selection.
	To address this gap in comparability, we perform a large-scale benchmark of more than 45 models for image classification, evaluating key efficiency aspects, including accuracy, speed, and memory usage.
	Our benchmark provides a standardized baseline for efficiency-oriented transformers.
	We analyze the results based on the Pareto front -- the boundary of optimal models.
	Surprisingly, despite claims of other models being more efficient, ViT remains Pareto optimal across multiple metrics.
	We observe that hybrid attention-CNN models exhibit remarkable inference memory- and parameter-efficiency.
	Moreover, our benchmark shows that using a larger model in general is more efficient than using higher resolution images.
	Thanks to our holistic evaluation, we provide a centralized resource for practitioners and researchers, facilitating informed decisions when selecting or developing efficient transformers.
	\footnote{\scriptsize\url{https://github.com/tobna/WhatTransformerToFavor}}
\end{abstract}

\section{Introduction}
\begin{figure}[t]
	\centering
	\includegraphics{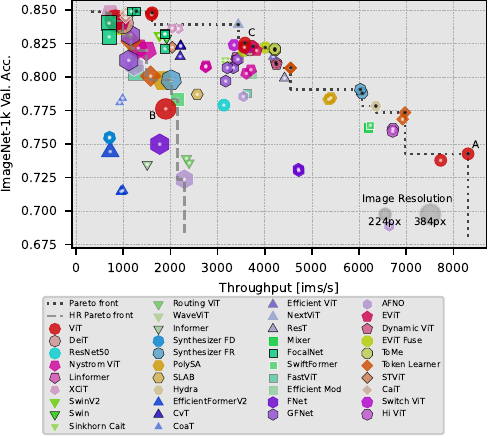}
	\caption{
		Pareto front (dotted line) showing the throughput-accuracy trade-off for efficient Vision Transformers.  Markers (varying in shape and hue) represent different efficiency strategies, with black dots highlighting Pareto optimal models. Marker size indicates fine-tuning resolution. Baselines ViT-Ti@224 (A), ViT-Ti@384 (B), and ViT-S@224 (C) are shown, along with the Pareto front for high-resolution images (dashed line).
		}
	\label{fig:throughput_vs_acc_size_imsize}
\end{figure}

The introduction of the Transformer~\cite{Vaswani2017} prompted its widespread adoption in both language and vision communities.
In particular for image classification, the Vision Transformer (ViT)~\cite{Dosovitskiy2021} has positioned itself as one of the best known applications of the original architecture, outperforming traditional CNN architectures on benchmarks like ImageNet \cite{Zhai2022,Yu2022}.
However, a major challenge in working with Transformer models is dealing with the computational complexity of the self-attention mechanism. 
This mechanism enables the Transformer to capture global dependencies between pairs of tokens, but it has a computational complexity of $\O(N^2)$ in the input length $N$, making it impractical for long sequences and high-resolution images.

Efforts have been made to reduce the computational load of Transformer models, 
particularly for resource-constrained settings, like embedded systems \cite{Wang2020a,Tabani2021}. 
Researchers have explored numerous strategies, like implementing sparse, local, or kernelized attention mechanisms. 
However, selecting the most efficient model that meets certain performance standards remains a challenging task, especially since ``\emph{efficiency}'' can refer to different concepts, like training resources, inference speed, or memory requirements.
Moreover, difficulties identifying the most efficient, yet highly performant vision transformers are accentuated by different training and evaluation conditions reported in the literature.
It is even more unclear how tradeoffs in efficiency for models proposed for NLP carry over to vision tasks.

To address these problems of fairly comparing efficient ViTs, we design an impartial test bed, training models from scratch on the same data, using the same setup.
While many papers at most report inference speed \cite{Patro2023,Liu2021,Ma2024,Vasu2023a}, our benchmark enables us to fairly provide empirical values across multiple dimensions of efficiency.
We identify common strategies employed to enhance model efficiency and utilize them to correlate changes in efficiency metrics with the respective strategies.
Through our benchmark, we provide a centralized resource of comprehensive baselines in the domain of image classification, conducting a thorough review of the current state of research on efficient Transformers.
In particular, we provide baselines for Transformers proposed for NLP, which can be trivially adapted to CV.

We analyze model speed and memory requirements both at training and inference time, as well as parameter efficiency.
This reveals that not all efficiency claims are realized.
Our analysis is based on the Pareto front, the set of models that provide an optimal tradeoff between model performance and one aspect of efficiency.
It lets us analyze the complex multidimensional tradeoffs involved in judging efficiency.
We plot inference throughput against accuracy in \Cref{fig:throughput_vs_acc_size_imsize} to visualize the Pareto front.
Here, a model is Pareto optimal (black dot) if and only if there is no model that is both more accurate and faster at the same time.
Through this analysis, we show that ViT remains Pareto optimal across three of the four empirical metrics we track, and also find other efficiency strategies that are Pareto optimal for different metrics.
This benchmark, based on more than 200 experiments, provides researchers and practitioners with a valuable resource, aiding them in choosing the most efficient and effective model architecture for their specific use case.

\subsection*{Main Results}
\begin{itemize}
	\item A fair \textbf{large-scale benchmark} for efficiency in transformer models run on \textbf{45+ models}, closing the comparability gap due to different experimental conditions.
	\item We find that a well-trained ViT still is Pareto optimal for three out of four metrics, \textbf{challenging claims} of other models being more efficient \cite{Xiong2021,Rao2021,Chang2023}.
	\item Our analysis uncovers \textbf{novel findings} like that in most cases, scaling up the model size is more efficient than scaling up the image resolution.
\end{itemize}

\section{Related Work}
This section provides an overview of relevant work on transformer surveys and approaches to measure efficiency, while \Cref{sec:taxonomy} delves into specific efficient architectures.

\paragraph{Surveys on Efficient Transformers}
Efficiency has become a critical aspect of transformers, leading to the creation of multiple surveys on efficient transformers providing valuable taxonomies and insights.
\cite{Tay2022} focuses on efficiency gains primarily in NLP, while \cite{Fournier2023} and \cite{Zhuang2023} collect general approaches for enhancing model efficiency, including those applicable to transformers.
\cite{Patro2023,Liu2023b} present an extensive list of efficient ViTs, classified based on aspects such as computational complexity, robustness, and transparency, comparing the efficiency in terms of parameters, ImageNet accuracy, and inference speed, using information from the original papers. 
Surveys on ViT-like models \cite{Han2023,Yang2022b,Khan2022} focus on categorizing models according to different vision tasks, while \cite{Zuo2022} specializes in dense prediction.
Specialized surveys investigate the application of transformers in specific domains, like action recognition \cite{Ulhaq2022}, image restoration \cite{Ali2023}, medical imaging \cite{Shamshad2023, Li2023b, He2023, Parvaiz2023}, remote sensing \cite{Aleissaee2023}, speech recognition \cite{Latif2023}, language processing \cite{Casola2022}, time series \cite{Wen2022}, or multimodal tasks \cite{Xu2023c}.

We build on this by analyzing efficiency gains in terms of broader strategies for efficiency.
In contrast to previous work, we close the gap of different training strategies and enable a fair comparison of efficient architectures under the same conditions based on more than new 200 experiments.

\paragraph*{Evaluation of Efficiency}
Efficiency evaluation in deep learning models is another area of investigation.
\cite{Bartoldson2023,Dehghani2022} provide an overview of efficiency aspects and metrics, along with measurement methodologies, highlighting the potential pitfalls of relying solely on theoretical metrics. 
\cite{Canziani2016} conducts a highly regarded survey on the efficiency of CNNs, and \cite{Liang2022a} compares the efficiency of their novel architectures with older models using the Pareto front of throughput and accuracy. 
Finally, \cite{Tay2021a} introduces a benchmark on synthetic data to quantify transformer model performance on long sequences in NLP, evaluating 10 models.

We extend previous work by evaluating more than 45 efficient Transformers for CV, measuring empirical metrics on the same hardware and using real data.
The in-depth analysis of our benchmark's results reveals new insights.

\section{Efficient Transformers for Vision}
We present our efficiency assessment methodology and briefly describe ViT, along with various efficient transformer variants.
These models from the domains of CV and NLP all claim to improve the efficiency of the baseline transformer.
We systematically select efficient transformers based on popularity, diversity, and novelty of their approaches.
See \Cref{apdx:model-selection} for more details on model selection.

\subsection{Quantifying Efficiency}

The term \emph{efficiency} can take on different meanings; 
it is therefore crucial to consider multiple dimensions when evaluating a model's efficiency. 
In this paper, we analyze model speed and memory requirements at training and inference time and contrast them with accuracy.

Due to the complex multidimensional tradeoffs, efficiency cannot be captured by a single number. 
Hence, we focus on the Pareto front -- the boundary of the landscape of efficient models -- to identify models that achieve the best trade-offs between two metrics. 
The Pareto front represents the set of models that take the best compromises between two metrics.
A Pareto optimal model outperforms every other model in at least one of the two metrics.
See the Pareto front in \Cref{fig:throughput_vs_acc_size_imsize}, where for each model that is not on the Pareto front (white dot) there is a model on the Pareto front (black dot) that is both faster and more accurate.

\subsection{Core Elements of ViT}
We briefly describe the key elements of ViT, that have been studied to make it more efficient, as well as its key bottleneck: the $\O(N^2)$ computational complexity of self-attention. 
By identifying these components, we can establish where and how the main strategies, that have been proposed to enhance transformer efficiency, change the architecture.
ViT is an adaptation of the original transformer model for image processing tasks.
Instead of text, it takes an image as input and converts it into a sequence of non-overlapping patches. Each patch is linearly embedded into a token of size $d$, with a positional encoding being added.
A classification token $[\textsc{cls}]$ is added to the sequence, which is then fed through a transformer encoder.
There, the self-attention mechanism computes the attention weights $A$ between tokens, utilizing the query (${Q \in \R^{N \times d}}$) and key ($K \in \R^{N \times d}$) matrices, calculated from the input sequence:
\begin{align*}
	A = \softmax \left( \frac{Q K^\top}{\sqrt{d_\text{head}}} \right) \in \R^{N \times N},
\end{align*}
where the softmax is calculated row-wise. 
This matrix encodes the global interactions between every possible pair of tokens.
It is also the reason why the attention mechanism has an inherent computational (space \& time) complexity of $\O(N^2)$.
The output sequence of the attention-mechanism is a weighted sum of the input, using the value matrix ($V$):
\begin{align}\label{eq:attention_mechanism}
	\begin{split}
		X_\text{out} = AV = \softmax \left( \frac{Q K^\top}{\sqrt{d_\text{head}}} \right) V.
	\end{split}
\end{align}
After self-attention, the sequence is passed through a feedforward network with MLP layers.
In the end, only the {$[\textsc{cls}]$} token is used for the classification decision.

\begin{figure*}[t]
	\centering
	\includegraphics[width=.65\textwidth]{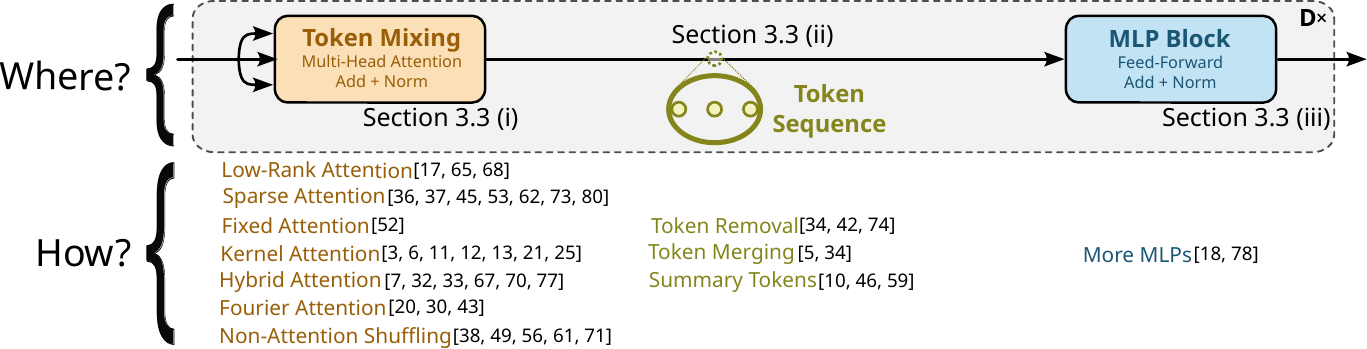}
	\caption{List of efficient Transformers (citation key in brackets) categorized at two levels: 1. Where does the approach change ViT? 2. How does the approach change ViT?}
	\label{fig:taxonomy}
\end{figure*}

\subsection{Efficiency-Improving Changes}
\label{sec:taxonomy}

After having gone over the backbone that constitutes a ViT, we discuss the most important modifications, which have been proposed to make it more efficient.
In order to analyze the overarching strategies that are taken, we classify the efficient models systematically using a two-step approach.
First, we identify the specific components — the token mixing mechanism, the token sequence, or the MLP block — of the baseline ViT architecture that are modified (i.e., “where”), as visualized in \Cref{fig:taxonomy}.
On the second level, we classify the different strategies to enhance the efficiency (i.e., “how”).
While this taxonomy is not meant to be a comprehensive overview of ViT-like models, it is proposed as a tool for identifying the best and most popular strategies to make vision transformers more efficient.
Note that our categorization scheme aligns closely with taxonomies that have been recently proposed \cite{Fournier2023,Tay2022,Selva2023}.

\paragraph*{(i) Token Mixing}
\hfill \\
The first and most popular approach is to change the \emph{token mixing} mechanism, which directly tackles the $\O(N^2)$ computational complexity of self-attention.
There are multiple strategies through which this can be accomplished.
Some methods approximate the attention mechanism with reduced computation, which can be achieved by matrix decomposition, changing the order of operations, or fixing attention values.
Other approaches combine attention with CNNs to perform sub-sampling in the attention mechanism or reduce the number of uses of the attention mechanism.
Finally, some methods discard the attention mechanism and instead introduce entirely new token mixing strategies.

\textbf{Low-rank attention} leverages the fact, that ${Q K^\top \in \R^{N \times N}}$ in \Cref{eq:attention_mechanism} is a matrix of rank $r \leq d \ll N$.
The \emph{Linformer} \cite{Wang2020} utilizes this to project the sequence direction of $K$ and $V$ down to dimension $k \ll N$, without reducing the informational content of the attention matrix too much.
Similarly, the \emph{Nyströmformer} \cite{Xiong2021} uses the Nyström method of matrix decomposition to approximate the matrix $Q K^\top$.
The approximate attention mechanism's output is then computed with linear complexity by applying the $\softmax$ to each part individually.
The approach of \emph{XCiT} \cite{ElNouby2021} utilizes a transposed attention mechanism:
\begin{align*}
	Y^\top = \softmax \left( \frac{1}{\sqrt d} Q^\top K\right) V^\top.
\end{align*}
Here, $Q^\top K \in \R^{d \times d}$ is used to replace the low-rank matrix $Q K^\top$ to define the globally informed filter
${\softmax \left( \frac{1}{\sqrt d} Q^\top K\right) \in \R^{d \times d}}$ which is applied to each token individually.
Since both $Q$ and $K$ are likely to be of rank $d$, the former most likely has full rank, which enables more efficient information encoding.

\textbf{Sparse attention} is an alternative approach to these dynamic approximations of global interactions.
Normal attention tends to focus on only a few input tokens, giving low scores to most others \cite{Kim2021}.
Sparse attention takes advantage of this by fixing most of the attention weights at zero and only calculating the most important ones.
The parts of the attention matrix, which are determined dynamically, can be based on a fixed pattern, like in the widely used \emph{Swin Transformer} \cite{Liu2021}, and \emph{SwinV2} \cite{Liu2022c}, where attention is performed only inside local sets of image tokens, or in \emph{HaloNet} \cite{Vaswani2021}, where each token can only attend to its neighbors.
Alternatively, \emph{Routing Transformer} \cite{Roy2021} determines sets of tokens that exchange information by grouping them based on their content.
In contrast, \emph{Sinkhorn Transformer} \cite{Tay2020} fixes local groups and lets tokens only attend to a different group using a dynamic permutation matrix.
\emph{Informer} \cite{Zhou2021} only calculates attention values for the $c \cdot \ln N$ most important tokens.
Another way of keeping the attention matrix sparse, is by sub-sampling the keys and values.
For example, \emph{Wave-ViT} \cite{Yao2022} uses a discrete wavelet transform for sub-sampling.

\textbf{Fixed attention} is an extreme example of setting attention values beforehand, which then only depend on the tokens positions.
This is explored in the \emph{Synthesizer} \cite{Tay2021}.

\textbf{Kernel attention} is an approach that changes the order of computations in \Cref{eq:attention_mechanism}. Instead of using the softmax on the product $QK^\top$, a kernel $\phi: \R^d \to \R^d$ is applied to the queries $Q$ and keys $K$ individually:
$$
	Y = \phi(Q) \phi(K)^\top V,
$$
which can be calculated with linear complexity $\O(N)$ in the number of tokens.
Various kernels have been proposed, including the random Gaussian kernel, introduced in \emph{Performer} \cite{Choromanski2021} and expanded on in the \emph{FourierLearner-Transformer} \cite{Choromanski2024} by adding a relative positional encoding in phase space.
\emph{Scatterbrain} \cite{Chen2021} combines this attention mechanism with sparse attention.
\emph{Poly-SA} \cite{Babiloni2023} uses the identity function $\phi \equiv \id$, \emph{Linear Transformer} \cite{Katharopoulos2020} uses an ELU, and \emph{SLAB} \cite{Guo2024} uses a ReLU kernel.
\emph{Hydra ViT} \cite{Bolya2022} uses an L2-normalization kernel and employs $H=d$ heads.

\textbf{Hybrid attention} combines convolutions with the attention mechanism.
\emph{EfficientFormerV2} \cite{Li2023d} uses convolutions initially to focus on local interactions, and then employs the attention mechanism to capture global interactions, while \emph{EfficientViT} \cite{Cai2023} utilizes convolutions first and kernel attention, using the ReLU kernel later on.
\emph{Next-ViT} \cite{Li2022b} alternates $N_\text{Conv} \in \N$ convolution blocks with a single attention block.
Another approach is to use convolutions inside the attention mechanism to create locally informed queries, keys, and values, as in \emph{CvT} \cite{Wu2021} and \emph{ResT} \cite{Zhang2021}, where it is used to subsample keys and values, and in \emph{CoaT} \cite{Xu2021}.

\textbf{Fourier attention} uses the Fast Fourier Transform (FFT) to reduce attention complexity to $\O(N \log N)$.
\emph{FNet} \cite{LeeThorp2022,Sevim2022} directly exploits the FFT for $\O(N \log N)$ complexity token mixing, while \emph{GFNet} \cite{Rao2021} utilizes it for a global convolution and \emph{AFNO} \cite{Guibas2022} uses an MLP in Fourier space.

\textbf{Non-attention shuffling} refers to techniques of capturing token interactions without using attention.
For example \emph{MLP-Mixer} \cite{Tolstikhin2021} uses a fully connected layer for global interactions, while \emph{FastViT} \cite{Vasu2023a} mixes tokens using depth-wise convolutions.
\emph{EfficientMod} \cite{Ma2024} modulates a convolutional context with a value matrix and
\emph{FocalNet} \cite{Yang2022a} modulates a hierarchy of contexts extracted by convolutions.
\emph{SwiftFormer} \cite{Shaker2023} calculates a global query to multiply with the keys and  also replaces the MLP block by a CNN.

\paragraph*{(ii) Token Sequence}
\hfill \\
The second category, \emph{token sequence}, is more prevalent in efficient transformers used in CV compared to NLP.
The idea is to remove redundant information typically contained in images, and in doing so, reducing computational costs without significantly affecting the model's performance.
Methods aim to reduce the token sequence by removing unnecessary image patches, such as the background, by merging tokens to minimize redundant information, or by summarizing the information into a smaller number of abstract tokens representing higher-level information.
These models leverage the $\O(N^2)$ complexity of the self-attention mechanism to attain a large reduction in computational cost, as removing $30\%$ of the tokens reduces the operations needed by around $50\%$.
While some of these methods can directly be applied to trained models, performance improves when implementing them during training \cite{Bolya2023}, which we do.

\textbf{Token Removal} selectively removes tokens while preserving critical information.
\emph{Dynamic ViT} \cite{Rao2021a} uses Gumbel-softmax for token retention probabilities, and \emph{A-ViT} \cite{Yin2022} learns halting probabilities to weight token outputs at different depths.
In contrast, \emph{EViT} \cite{Liang2022a} avoids introducing extra parameters by utilizing the previous layer's attention matrix.

\textbf{Token merging} is another strategy to remove redundant information.
While a version of \emph{EViT} \cite{Liang2022a} merges the unimportant tokens, \emph{ToMe} \cite{Bolya2023} merges tokens based on their similarity, using a fast bipartite matching algorithm.

\textbf{Summary tokens} condense the token sequence into a few new tokens, unlike the previous methods that modify existing tokens.
\emph{CaiT} \cite{Touvron2021a} uses cross-attention on a single token to gather global information on the classification decision in the last layers.
\emph{Token Learner} \cite{Ryoo2021} creates a set number of summary tokens by employing dynamic sums over the image tokens, and \emph{STViT} \cite{Chang2023} initializes summary tokens with local information by using strided convolutions, then injecting global information into those using cross-attention.

\paragraph*{(iii) MLP Block}
\hfill \\
The final way in which proposed methods change the architecture of transformers is by moving compute to the MLP block, which has linear complexity with respect to the sequence length.
Despite room for efficiency gains, we have identified only two methods taking this approach, which we include to make this benchmark more diverse.

\textbf{More MLPs} can be used to move compute into these efficient blocks.
\emph{Switch Transformer} \cite{Fedus2022} introduces multiple parameter sets for different sets of tokens, increasing the number of parameters without significantly introducing extra computations.
In contrast, \emph{HiViT} \cite{Zhang2023c} exchanges early attention layers for additional MLP blocks.

Out categorization offers a structured framework for understanding approaches of improving ViT efficiency.
This enables us to compare broad trends across strategies.

\section{Experimental Design}
We conduct an extensive series of over 200 experiments on the more than 45 models described in \Cref{sec:taxonomy} to evaluate which models are ideal under a given set of constraints.
We measure efficiency w.r.t. the full models rather than individual attention modules, since methods that reduce the token sequence (\Cref{sec:taxonomy} (ii)), and others, like XCiT, require additional modifications to the architecture, which we include.
Note, however, that many models only introduce changes in the attention mechanism, effectively turning our benchmark into a module-level comparison for these.

\subsection{Baselines}
To find the most efficient models, we need to quantify their efficiency gains by comparing them to the original ViT \cite{Dosovitskiy2021} and its follow-up versions \cite{Touvron2021b,Touvron2022}.
We also include ResNet50 \cite{He2016} in our evaluation as a representative baseline for CNN architectures and a point of reference across papers.
All evaluations are based on the ImageNet-1k dataset for classification \cite{Deng2009} as it is one of the most commonly used datasets in this domain.
Results on other datasets
(Stanford Cars \cite{Krause2013}, Oxford Flowers 102 \cite{Nilsback2008}, MIT Places365 \cite{Zhou2014})
can be found in \Cref{apdx:other-datasets}.

\subsection{Training Pipeline}
We compare models on even grounds by training with a standardized pipeline.
To reduce bias, our pipeline is relatively simple and only consists of elements commonly used in CV.
In particular, we refrain from using knowledge distillation to prevent introducing bias from the choice of teacher model.
Any orthogonal techniques, like quantization, sample selection, and others, are not included as they can be applied to every model and would manifest as a systematic offset in the results.
To avoid confounding effects from limited training data, we pre-train all models on ImageNet-21k \cite{Ridnik2021}.
While data scarcity impacts performance, we focus on evaluating each architecture's maximum potential.

Models are trained for a total of 140 epochs using the pipeline introduced in \cite{Touvron2022}. This is an updated version of DeiT \cite{Touvron2021b}, which has been adopted throughout the literature for training efficient ViTs (see \Cref{table:pipeline_comparisons}).
As this pipeline only employs standard CV practices, we consider it a fair point of comparison for evaluating all models.

Pretraining is conducted on ImageNet-21k for 90 epochs at resolutions of $224$ and $192$ pixels, followed by fine-tuning on ImageNet-1k at $224$ and $384$ pixels for 50 epochs.
In cases where model training is unstable, we default back to the hyperparameters reported in the corresponding publications.
All training is conducted using 4 or 8 NVIDIA A100 GPUs.
While most of the models work well with our training pipeline, some did not converge.
See \Cref{apdx:training-pipeline} for more details.

\subsection{Efficiency Metrics}
As we saw, efficiency has multiple dimensions, and therefore we need to capture a variety of metrics.
For comparison with other papers and to evaluate their use as proxy metrics for speed and memory requirements, we capture the theoretical metrics of number of parameters and FLOPS.
These are used by the community to provide estimates of the model's representational capacity and computational requirements \cite{Patro2023,Tay2022,Rao2021}.
However, theoretical metrics do not always correlate with real-world performance, especially when comparing different architectures \cite{Dehghani2022,Bartoldson2023}.
Instead, empirical metrics, obtained by running models on hardware, offer more informative evaluations, which is why we track GPU time for training, inference throughput at optimal batch sizes, and VRAM requirements.
While these empirical metrics can be sensitive to hardware and software configurations, we ensure consistent evaluations by employing the same setup for all models.
We evaluate the variance of our metrics in \Cref{apdx:metric-variation}.

\section{Results}
\begin{table}[t]
	\centering
	\caption{ImageNet-1k accuracy of the original papers and the new training pipeline. \checkmark marks original pipelines based on DeiT \cite{Touvron2021b}. Models are trained on $224$px images, unless marked with $\uparrow$ ($384$px) or $\downarrow$ ($112$px). Results with ${}^D$ use knowledge distillation and ${}^E$ uses exponential moving average.}
	\label{table:pipeline_comparisons}
	\resizebox{.35\textwidth}{!}{
		\begin{tabular}[t]{lclc}
			\toprule
			\multirow{2}{*}{Model} & \multicolumn{2}{c}{Original} & Ours                                     \\
			                       & DeiT                         & Accuracy                & Accuracy       \\
			\cmidrule(r){1-1} \cmidrule(rl){2-3} \cmidrule(l){4-4}
			ViT-S (DeiT)           & \checkmark                   & 79.8                    & \textbf{82.54} \\
			ViT-S (DeiT III)       &                              & 82.6                    & 82.54          \\
			XCiT-S                 & \checkmark                   & 82.0                    & \textbf{83.65} \\
			Swin-S                 & \checkmark                   & 83.0                    & \textbf{84.87} \\
			SwinV2-Ti              &                              & 81.7                    & \textbf{83.09} \\
			Wave-ViT-S             &                              & 82.7                    & \textbf{83.61} \\
			Poly-SA-ViT-S          &                              & 71.48                   & \textbf{78.34} \\
			SLAB-S                 & \checkmark                   & \textbf{80.0}           & 78.70          \\
			EfficientFormer-V2-S0  &                              & \textbf{75.7}${}^D$     & 71.53          \\
			CvT-13                 &                              & \textbf{83.3}$\uparrow$ & 82.35          \\
			CoaT-Ti                & \checkmark                   & 78.37                   & 78.42          \\
			EfficientViT-B2        &                              & \textbf{82.7}$\uparrow$ & 81.52          \\
			NextViT-S              &                              & 82.5                    & \textbf{83.92} \\
			ResT-S                 & \checkmark                   & 79.6                    & \textbf{79.92} \\
			FocalNet-S             &                              & 83.4                    & \textbf{84.91} \\
			SwiftFormer-S          &                              & \textbf{78.5}${}^D$     & 76.41          \\
			FastViT-S12            & \checkmark                   & \textbf{79.8}$\uparrow$ & 78.77          \\
			EfficientMod-S         & \checkmark                   & \textbf{81.0}           & 80.21          \\
			GFNet-S                &                              & 80.0                    & \textbf{81.33} \\
			EViT                   & \checkmark                   & 79.4                    & \textbf{82.29} \\
			DynamicViT-S           &                              & \textbf{83.0}${}^D$     & 81.09          \\
			EViT Fuse              & \checkmark                   & 79.5                    & \textbf{81.96} \\
			ToMe-ViT-S             & \checkmark                   & 79.42                   & \textbf{82.11} \\
			TokenLearner-ViT-8     &                              & 77.87$\downarrow$       & \textbf{80.66} \\
			STViT-Swin-Ti          & \checkmark                   & 80.8                    & \textbf{82.22} \\
			CaiT-S24               & \checkmark                   & 82.7                    & \textbf{84.91} \\
			\bottomrule
		\end{tabular}}
\end{table}

\subsection{Improved Training Pipeline}

To validate the fairness of our training pipeline, in \Cref{table:pipeline_comparisons} we compare ImageNet-1k accuracy reported in the original papers (whenever reported) with the one we obtained to see if there are any obvious outliers.
Note, that a substantial fraction (13 out of 26) of papers explicitly base their training pipelines on DeiT \cite{Touvron2021b}, making them a good fit for training with the updated version in DeiT III \cite{Touvron2022}.
We observe that almost all models trained with this pipeline achieve a higher accuracy; 0.85\% on average, with the largest gain being +6.86\% for Poly-SA.
The models reporting higher performance than what we obtain use knowledge distillation, which we avoid to not induce bias, or a higher resolution, which we find is inefficient.

Most categories contain models that achieve peak accuracy of approximately 85\% (see \Cref{apdx:acc-per-class}). However, two notable exceptions are observed: the \emph{Kernel Attention} class, which might pose challenges in optimizing models, and the \emph{Fixed Attention} class, where the use of a constant attention matrix inherently results in a lower accuracy.
Overall, our selection of the training pipeline as a reference for a fair comparison is supported by the improvements in results compared to the original papers, regardless of architectural differences.

\begin{figure}[t]
	\centering
	\includegraphics{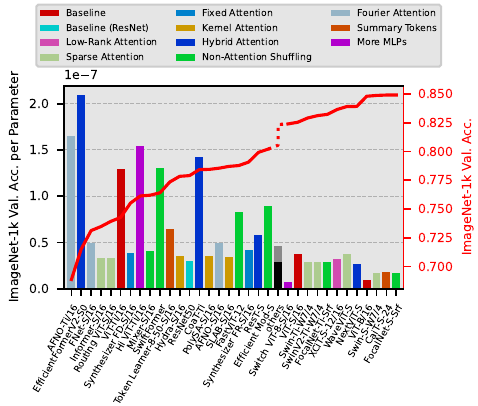}
	\caption{Accuracy (red line, right y-axis) and accuracy per parameter (bars, left y-axis) of models ordered by accuracy at a resolution of $224$px. 14 of 47 models of intermediate size are grouped into \emph{others}. See \Cref{apdx:acc-per-param} for the full plot.}
	\label{fig:model_vs_acc_per_param_and_acc_excerpt}
\end{figure}

\subsection{Number of Parameters}
The number of parameters is used in the literature as a proxy metric for tracking the model complexity and the overall computational cost of using a model.
When analyzing the parameter efficiency of different models (\Cref{fig:model_vs_acc_per_param_and_acc_excerpt}), it is evident that for most smaller models, the accuracy per parameter remains relatively constant at a baseline level of about $4 \times 10^{-8} \frac{\%}{\text{param.}}$.
This value is approximately halved for larger models, indicating diminishing returns on scaling the model size.
For ViT, there is a noticeable drop in accuracy per parameter across the model sizes.
While ViT-Ti outperforms models of comparable accuracy, and ViT-S performs on par with similar models, ViT-B slightly underperforms when compared to other, larger models.
Significant outliers are the smaller \emph{Hybrid Attention} models EfficientFormerV2-S0 and CoaT-Ti, which exhibit the highest accuracy per parameter, as well as those \emph{Non-attention Shuffling} models which incorporate convolutions, like SwiftFormer and FastViT.
In the \emph{MLP} category, we find HiViT to have significantly higher and Switch ViT with significantly lower accuracy per parameter than surrounding models.
This suggests that the combination of attention and convolutions allows for the development of highly parameter-efficient models.

\subsection{Speed}
\begin{figure}[t]
	\centering
	\includegraphics{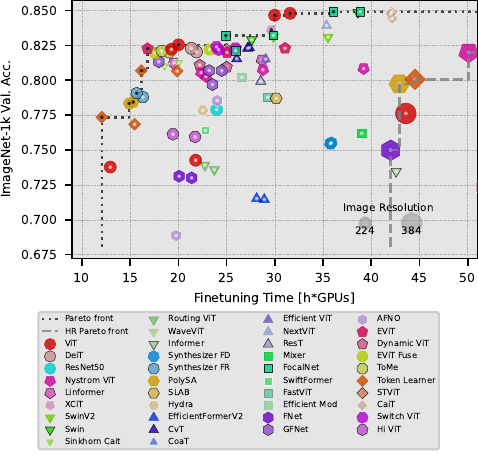}
	\caption{Pareto front of finetuning time and accuracy for models which need less than 50 hours for finetuning. We include the full plot in \Cref{apdx:finetuning-time}.}
	\label{fig:finetuning_time_vs_acc_size_prop_imsize_excerpt}
\end{figure}
Whether driven by strict requirements for real-time processing or the desire to obtain model outputs within reasonable timeframes, inference speed directly impacts the usability of deployed models.
The models we evaluate often claim a superior throughput vs. accuracy trade-off compared to ViT.
However, our evaluation in \Cref{fig:throughput_vs_acc_size_imsize} reveals that ViT remains Pareto optimal at all model sizes.
Only few models, namely Synthesizer-FR, NextViT, and some \emph{Sequence Reduction} models, show improvements in the Pareto front when compared to a ViT of comparable size.

We also analyze the Pareto optimality with models finetuned on Stanford Cars \cite{Krause2013}, Oxford Flowers 102 \cite{Nilsback2008}, and MIT Places365 \cite{Zhou2014} in \Cref{apdx:other-datasets}, observing a high Spearman correlation between the accuracy of different datasets.
In particular, the Pareto optimal models of throughput and accuracy are largely the same, which is a strong indicator for the generalizability of our results.
Even when assessing throughput on a CPU instead of a GPU, we observe only a high Spearman correlation with only few outliers.
Again, we find the Pareto front to be robust to the choice of hardware.
See \Cref{apdx:speed-hardware} for more information about these results.

In the Pareto front of fine-tuning time in \Cref{fig:finetuning_time_vs_acc_size_prop_imsize_excerpt}, we see many similarities to the one of inference time.
Here, some of the \emph{token sequence} based models are highly efficient.
In particular, TokenLearner emerges as a standout performer with the fastest fine-tuning speed while achieving a competitive accuracy of 77.35\%.
For larger models, we find ViT-B and FocalNet (-Ti and -S) to be Pareto optimal.

Generally, ViT is still a solid choice for speed, with some \emph{Token Sequence} models trading off accuracy for speed.

\begin{figure*}[t]
	\centering
	\includegraphics{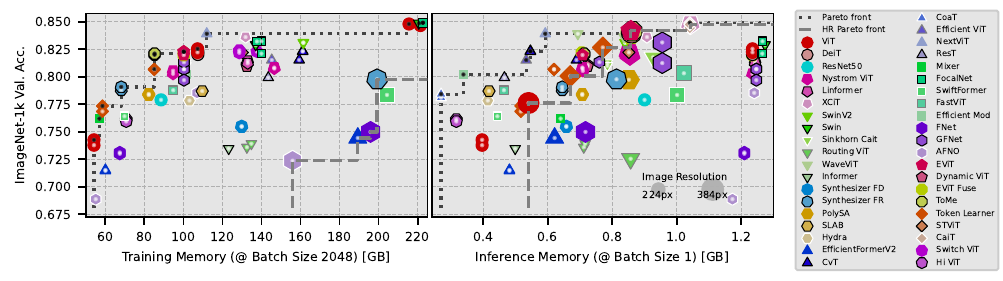}
	\caption{Pareto front (dotted line) of training memory at our default batch size of $2048$ (left) and inference memory at the minimum batch size of $1$ (right) and accuracy for models with less than 225GB of VRAM for training and 1.25GB for inference.}
	\label{fig:memory_vs_acc_size_imsize}
\end{figure*}

\subsection{Memory}
\Cref{fig:memory_vs_acc_size_imsize} (right) reveals that the tradeoffs when optimizing for VRAM usage during inference are very different from the ones for speed or training VRAM (left side), making it the only metric where ViT is \emph{not} Pareto optimal.
At inference time, the \emph{Hybrid Attention} models CoaT, CvT, ResT, and NextViT stand out from the rest.
Additionally, CaiT, and EViT@384px for the larger sizes and EfficientMod for the smaller ones are Pareto optimal.
This observation suggests that, similar to the number of parameters, \emph{Hybrid Attention} models excel in low memory environments.

In contrast, the aspect of training memory (\Cref{fig:memory_vs_acc_size_imsize}; left) exhibits a similar pattern as observed in throughput.
Here, CoaT, CvT, and ResT need more memory than other models of comparable accuracy and again ViT is Pareto optimal, along with some \emph{Sequence Reduction} models and Synthesizer-FR.
NextViT is the only model that is Pareto optimal for both training and inference memory.
When training at a resolution of $384$px most models need more than $225$GB of VRAM and therefore are not plotted.

Overall, inference memory is the most different from the other Pareto fronts with \emph{Hybrid Attention} models excelling.

\subsection{Scaling Behaviors}
Our observations reveal that fine-tuning at a higher resolution is inefficient. %
While it may result in improved accuracy, it entails a significant increase in computational cost, leading to a substantial reduction in throughput.
In turn, scaling up the model ends up being more efficient.
This is shown in \Cref{fig:throughput_vs_acc_size_imsize}, where the best models using high resolution images are two to four times slower than a low resolution model at the same accuracy (compare \emph{Pareto front} and \emph{HR Pareto front}).
Consequently, no model fine-tuned at the higher resolution is Pareto optimal.
For example, when scaling from tiny to small (A to C in \Cref{fig:throughput_vs_acc_size_imsize}), we gain a lot more accuracy for the lost speed than when scaling up the image size (A to B).
For training memory, this trend is even more pronounced (see \emph{HR Pareto front} in \Cref{fig:memory_vs_acc_size_imsize}), with most of the high resolution models requiring more than $225$GB of VRAM for training and thus being off the chart.
Overall, many papers focus on reducing the $\O(N^2)$ complexity of self-attention to deal with long sequences, however sequences in image classification are typically not long enough to show the theoretical gains.
On the contrary, it is more efficient to use shorter sequences instead.
We believe that efficient mechanisms focusing on these shorter sequences are a fruitful area for future research.

\subsection{Correlation of Metrics}
\begin{table}[t]
	\centering
	\caption{Correlation between the number of floating point operations, number of parameters ($\theta$), fine-tuning time, training memory, inference time, and inference memory.}
	\label{table:correlation_of_metrics}

	\resizebox{.35\textwidth}{!}{
		\begin{tabular}{lcccccc}
			\toprule
			\multirow{2}{*}{$\operatorname{corr}(x, y)$} & \multirow{2}{*}{$\theta$} & \multicolumn{2}{c}{Training} & \multicolumn{2}{c}{Inference}                        \\
			                                             &                           & Time                         & Mem                           & Time          & Mem  \\
			\cmidrule(r){1-1} \cmidrule(lr){2-2} \cmidrule(lr){3-4} \cmidrule(l){5-6}
			FLOPS                                        & 0.30                      & 0.72                         & \textbf{0.85}                 & 0.48          & 0.42 \\
			\textit{$\theta$}                            &                           & 0.05                         & 0.18                          & 0.02          & 0.40 \\
			Training Time                                &                           &                              & \textbf{0.89}                 & \textbf{0.81} & 0.17 \\
			Training Mem                                 &                           &                              &                               & 0.71          & 0.48 \\
			Inference Time                               &                           &                              &                               &               & 0.13 \\
			\bottomrule
		\end{tabular}}
\end{table}

We run a correlation analysis between the efficiency metrics in \Cref{table:correlation_of_metrics}.
The highest correlation of $0.89$ is between fine-tuning time and training memory. This suggests a common underlying factor or bottleneck, possibly related to the necessity of memory reads during training.
Understanding this relationship can provide valuable insights into the factors influencing training efficiency.
Intriguingly, the highest correlation coefficient involving a theoretical metric is $0.85$ between FLOPS and training memory, suggesting the possibility for a rough estimation.
For other metrics, we reaffirm the limited reliability of estimating computational costs solely based on theoretical metrics \cite{Dehghani2022,Bartoldson2023}.
Consequently, assessing model efficiency in practice requires the empirical measurement of throughput and memory requirements.

\section{Discussion \& Conclusion}
\label{sec:conclusion}

\begin{figure}[t]
	\centering
	\resizebox{\columnwidth}{!}{
		\begin{tikzpicture}
			\node[align=center] (WTF) at (0, 0) {Which\\Transformer\\to Favor?};
			\node[align=center] (Const) at (2.8, 1.5) {1. \textbf{Main}\\\textbf{Constraint}};
			\node[align=center] (Phase) at (4.9, 1.5) {2. \textbf{Phase}};
			\node[align=center,anchor=west] (A) at (6.3, 1.5) {3. \textbf{Transformers to Favor}};

			\node[anchor=west] (Speed) at (2.2, .55) {Speed};
			\node[anchor=west] (Mem) at (2.2, -.55) {Memory};
			\node[anchor=west] (SpInf) at (4.2, .8) {Inference};
			\node[anchor=west] (SpTrain) at (4.2, .3) {Training};
			\node[anchor=west] (MemInf) at (4.2, -.3) {Inference};
			\node[anchor=west] (MemTrain) at (4.2, -.8) {Training};

			\draw[->,out=0,in=180] (WTF) to (Speed);
			\draw[->,out=0,in=180] (WTF) to (Mem);
			\draw[->,out=0,in=180] (Speed) to (SpInf);
			\draw[->,out=0,in=180] (Speed) to (SpTrain);
			\draw[->,out=0,in=180] (Mem) to (MemInf);
			\draw[->,out=0,in=180] (Mem) to (MemTrain);

			\node[anchor=west] (ASI) at (6.3, .8) {\{\textcolor{red}{ViT} \textcolor{blue}{NextViT}, \textcolor{violet}{ToMe}, \textcolor{violet}{TokenLearner}, \textcolor{violet}{...}\}};
			\node[anchor=west] (AST) at (6.3, .3) {\{\textcolor{red}{ViT}, \textcolor{violet}{TokenLearner}, \textcolor{orange}{FocalNet}\}};
			\node[anchor=west] (AMI) at (6.3, -.3) {\{\textcolor{red}{ViT}, \textcolor{blue}{NextViT}, \textcolor{blue}{CoaT}, \textcolor{blue}{CvT}, \textcolor{blue}{...}\}};
			\node[anchor=west] (AMT) at (6.3, -.8) {\{\textcolor{red}{ViT}, \textcolor{violet}{ToMe}, \textcolor{violet}{TokenLearner}, \textcolor{blue}{NextViT}, ...\}};

			\draw[->] (SpInf) to (ASI);
			\draw[->] (SpTrain) to (AST);
			\draw[->] (MemInf) to (AMI);
			\draw[->] (MemTrain) to (AMT);
		\end{tikzpicture}}
	\caption{Suggestions on \emph{Which Transformer to Favor} in image classification.
		First, choose the main constraint (1.), speed, or memory, and the phase you want to optimize this constraint at (2.): Inference or training.
		Based on that, we provide recommendations on which transformer to favor (3.) from the taxonomy classes of \textcolor{red}{Baseline}, \textcolor{violet}{Sequence Reduction}, \textcolor{blue}{Hybrid Attention}, \textcolor{orange}{Non-Attention Shuffling} and \textcolor{green}{Sparse Attention}.
		See \Cref{apdx:pareto-optimal-models} for a full list of Pareto optimal models for each metric.}
	\label{fig:wtf_answer}
\end{figure}

In this paper, we provide a thorough benchmark of efficiency oriented transformers for vision, facilitating a fair, competitive comparison of contributions from NLP and CV and revealing some unexpected trends.
Our analysis shows ViT's continued Pareto optimality across multiple metrics, reaffirming its effectiveness as a baseline, despite other models claiming to be more efficient.
Results also prove that scaling the model size is, in most cases, more efficient than using higher resolution images.
This goes against the trend of efficient models being evaluated using high resolution images \cite{Touvron2021a,Liu2021,ElNouby2021,Vasu2023a}.
Comparing different efficiency metrics, we reaffirm the limitations of estimating computational costs solely based on theoretical metrics.

When addressing the question of \emph{which transformer to favor},
our benchmark offers actionable insights in the form of models and strategies to use (see \Cref{fig:wtf_answer}).
ViT remains the preferred choice overall. However, \emph{Token Sequence} methods can become viable alternatives when speed and training efficiency are of importance.
For scenarios with significant inference memory constraints, considering \emph{Hybrid} CNN-attention models can prove advantageous.

Looking ahead, we posit that the establishment of this comparable, fair benchmark and analysis will drive substantial advancements in efficiency-oriented transformers for CV, benefitting researchers and practitioners making architectural decisions for resource-constrained environments.

\section*{Acknowledgements}

This work was funded by the Carl-Zeiss Foundation under the Sustainable Embedded AI project (P2021-02-009), by the EU project SustainML (Horizon Europe grant agreement No 101070408), and by the Albatross project (01IW24002).
All compute was done thanks to the Pegasus cluster at DFKI.

{\small
	\bibliographystyle{ieee_fullname}
	\bibliography{main_bib}
}

\newpage
\onecolumn
\appendix

\section{Implementational Details}
\subsection{Efficient Transformers}
Our implementation is based on the \inlinecode{PyTorch} framework \cite{Paszke2019} and the \inlinecode{timm} package \cite{Wightman2019}.
Whenever possible, we utilize the original implementations provided by the authors of each transformer variant.
In cases where PyTorch implementations were not available, we adopt other existing open-source implementations or develop our own as needed.
Specific sources of each implementation can be found in the corresponding code files or the README file\footnote{\url{https://github.com/tobna/WhatTransformerToFavor}}.
We make minimal modifications to these implementations to ensure consistent handling of input and output sizes and for the accommodation of additional loss terms.
For models that follow the overarching structure of ViT \cite{Dosovitskiy2021}, we only change the implementation of the attention mechanism, keeping the rest the same as ViT.

\subsection{Training Pipeline Details}
\label{apdx:training-pipeline}

\begin{table}[ht!]
    \centering
    \renewcommand{\arraystretch}{0.8}
    \caption{Training pipeline details and hyperparameters. In 3-Augment, we always choose one of grayscale, solarize, or Gaussian blur at random.}
    \label{tab:pipeline_details}
    \small
    \begin{tabular}{lcc}
        \toprule
                              & Pretrain                                                       & Finetune                                  \\
        \cmidrule(r){1-1} \cmidrule(l){2-3}
        Dataset               & \makecell[c]{ImageNet-21k \cite{Ridnik2021}}                   & \makecell[c]{ImageNet-1k \cite{Deng2009}} \\
        Epochs                & 90                                                             & 50                                        \\
        LR                    & $3 \times 10^{-3}$                                             & $3 \times 10^{-4}$                        \\
        Schedule              & \multicolumn{2}{c}{cosine decay}                                                                           \\
        Min LR                & \multicolumn{2}{c}{$10^{-5}$}                                                                              \\
        Batch Size            & \multicolumn{2}{c}{$2048$}                                                                                 \\
        Warmup                & \multicolumn{2}{c}{linear}                                                                                 \\
        Warmup Epochs         & \multicolumn{2}{c}{$5$}                                                                                    \\
        Warmup LR             & \multicolumn{2}{c}{$10^{-6}$}                                                                              \\
        Weight decay          & \multicolumn{2}{c}{$0.02$}                                                                                 \\
        Grad Clipping         & \multicolumn{2}{c}{$1.0$}                                                                                  \\
        Label Smoothing       & \multicolumn{2}{c}{$0.1$}                                                                                  \\
        Drop Path Rate        & \multicolumn{2}{c}{$0.05$}                                                                                 \\
        Optimizer             & \multicolumn{2}{c}{\makecell[c]{Lamb \cite{You2020}}}                                                      \\
        Optimizer $\eps$      & \multicolumn{2}{c}{$10^{-7}$}                                                                              \\
        LayerScale Init       & \multicolumn{2}{c}{$10^{-4}$}                                                                              \\
        Dropout prob.         & \multicolumn{2}{c}{$0.0$}                                                                                  \\
        Mixed precision       & \multicolumn{2}{c}{AMP}                                                                                    \\

        \cmidrule(r){1-1} \cmidrule(l){2-3}
        \textbf{Augmentation} & \multicolumn{2}{c}{\makecell[c]{3-Augment \cite{Touvron2022}}}                                             \\
        Flip prob.            & \multicolumn{2}{c}{$0.5$}                                                                                  \\
        Crop                  & \multicolumn{2}{c}{Simple Random Crop}                                                                     \\
        Resize to             & \makecell[c]{$224 \times 224$ or                                                                           \\ $192 \times 192$} & \makecell[c]{$224 \times 224$ or \\ $384 \times 384$} \\
        Grayscale prob.       & \multicolumn{2}{c}{$0.33$}                                                                                 \\
        Solarize prob.        & \multicolumn{2}{c}{$0.33$}                                                                                 \\
        Gauss. Blur prob.     & \multicolumn{2}{c}{$0.33$}                                                                                 \\
        Color Jitter Fact.    & \multicolumn{2}{c}{$0.3$}                                                                                  \\
        Cutmix prob.          & \multicolumn{2}{c}{$1.0$}                                                                                  \\
        Normalization         & \multicolumn{2}{c}{\makecell[c]{\begin{tabular}{llll}
                                                                            $\mu =$    & $(0.485,$ & $0.456,$ & $0.406)$ \\
                                                                            $\sigma =$ & $(0.229,$ & $0.224,$ & $0.225)$
                                                                        \end{tabular}}}                               \\
        \cmidrule(r){1-1} \cmidrule(l){2-3}
        GPUs                  & \multicolumn{2}{c}{4 (or 8) NVIDIA A100}                                                                   \\
        \bottomrule
    \end{tabular}
\end{table}

The training pipeline is taken from \emph{DeiT III} \cite{Touvron2022}, building upon the well-established pipeline outlined in \emph{DeiT} \cite{Touvron2021b}.
\Cref{tab:pipeline_details} presents the default hyperparameters and training pipeline details.
Our training process involves initial pretraining at resolutions of $192 \times 192$ and $224 \times 224$ pixels, followed by fine-tuning at $224 \times 224$ pixels.
The models pretrained at $224 \times 224$ pixels additionally get finetuned at $384 \times 384$ pixels, as long as this fits onto 8 NVIDIA-A100s (80 GB).
In cases where training (especially pretraining) is unstable, we adjust some hyperparameters in accordance with the original papers, beginning with the learning rate, warmup epochs, and weight decay.

\subsection{Efficiency Evaluation}
For calculating the theoretical metrics, we directly count the number of parameters of the PyTorch implementation by iterating over \inlinecode{model.parameters()}. We count the number of floating-point operations using the \inlinecode{fvcore} package\footnote{\url{https://github.com/facebookresearch/fvcore/tree/main}}, with additional handles for counting FLOPs from operations beyond matrix multiplication.

Memory requirements, both at training and inference time, are measured using \inlinecode{torch.cuda.max\_memory\_allocated()}.
To measure training time, we capture differences of Python's \inlinecode{time.time} across each training epoch.
Throughput is measured using CUDA Events\footnote{\url{https://pytorch.org/docs/stable/generated/torch.cuda.Event.html}} and averaged over $1000$ iterations.
This is additionally done at multiple batch sizes, including powers of two, and others close to the GPUs VRAM limit. We report the best result together with the corresponding batch size.

\subsection{Variability of Efficiency Metrics}
\label{apdx:metric-variation}
\begin{figure*}[ht!]
    \centering
    \resizebox{.8\textwidth}{!}{\includegraphics{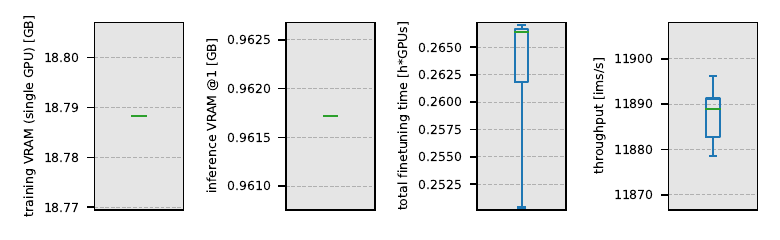}}
    \caption{Box plots of efficiency metric ranges over $8$ independent training runs.}
    \label{fig:variation_boxplots}
\end{figure*}

To validate the reliability of our measurements of these metrics, we inspect each metric's variability by empirically assessing it for the baseline of ViT.
\Cref{fig:variation_boxplots} shows ranges of the empirical efficiency metrics across $8$ training runs.
Particularly, we observe negligible variability in the training and inference VRAM measurements, with only marginal variability in time-based metrics.
For instance, the range of throughput measured is $11878$ to $11896$ images per second, equivalent to $0.15\%$ of the mean, with a standard deviation of $6.48$, $0.05\%$ of the mean.
Finetuning time measurements have a slightly larger range from $0.2504$ to $0.2670$ GPU-hours per epoch, or $6.33\%$ of the mean, with a standard deviation of $0.0075$, which represents $2.84\%$ of the mean.
We attribute this marginal increase in variation predominantly to external factors, such as server temperature and load, as well as nuanced dissimilarities between specific instances of the same hardware.
Overall, the measurements of all efficiency metrics are very robust.

\section{Model Selection}
\label{apdx:model-selection}
We systematically select efficient transformers based on diversity, popularity, and novelty of their approaches.
These models, from the domains of CV and NLP, all claim to improve the efficiency of the baseline transformer.
We select papers that are listed under the term \emph{efficient transformer}\footnote{\url{https://www.semanticscholar.org/search?year[0]=2018&year[1]=2024&q=efficient\%20transformer&sort=total-citations}} and are either published since 2018 with atleast 50 citations or published since 2023 with atleast 10 citations.
We also select papers that are listed under \emph{efficient ViT}\footnote{\url{https://www.semanticscholar.org/search?year[0]=2018&year[1]=2024&q=efficient\%20ViT&sort=total-citations}} since 2018 with atleast 5 citations.
Additionally, we include papers from recent high-ranking conferences, like NeurIPS 2023, AAAI 2024, ICML 2024, ICLR 2024, and WACV 2024.
We then add models that divise unusual approaches to make transformers more efficient, like the \emph{Switch Transformer} or \emph{FNet}, to increase the diversity of the benchmarked strategies.
We only keep papers that introduce changes to the transformer architecture in order to make it more efficient and have also published their code\footnote{Full PyTorch implementation needed to be published until the 1st of July 2024.}.
We exclude models that have architectures specifically designed for one specific task only, like semantic segmentation, point clouds, or spherical images.

\section{Other Datasets \& Tasks}
\label{apdx:other-datasets}
\begin{table}[ht!]
    \centering
    \caption{Hyperparameter changes for finetuning models on other datasets. All the other parameters are the same as in \Cref{tab:pipeline_details}.}
    \label{tab:other_dataset_hyperparameters}
    \renewcommand{\arraystretch}{1.}
    \begin{tabular}{lccc}
        \toprule
        Parameter  & Cars & Flowers & Places \\
        \cmidrule(r){1-1} \cmidrule(l){2-4}
        epochs     & 2000 & 2000    & 50     \\
        lr         & 3e-4 & 3e-4    & 3e-4   \\
        batch size & 1024 & 256     & 2048   \\
        resize to  & 224  & 224     & 224    \\
        \bottomrule
    \end{tabular}
\end{table}

\begin{table*}[ht!]
    \centering
    \caption{Model performance on different datasets and correlation coefficients of datasets.}
    \small
    \label{tab:other_datasets}
    \renewcommand{\arraystretch}{1.}
    \begin{tabular}{llcccc}
\toprule
	Model & Taxonomy Class & ImageNet-1k & Cars & Flowers & Places \\
	\cmidrule(r){1-2} \cmidrule(l){3-6}
	 ViT-S/16 & Baseline & 82.54 & 91.58 & 98.87 & 58.56 \\
	 ViT-Ti/16 & Baseline & 74.27 & 90.32 & 96.66 & 55.14 \\
	 ViT-B/16 & Baseline & 84.80 & 92.00 & 99.53 & 59.13 \\
	 DeiT-S/16 & Baseline & 82.29 & 92.24 & 98.89 & 58.08 \\
	 ResNet50 & Baseline (ResNet) & 77.90 & 92.63 & 96.88 & 56.06 \\
	 Nystrom ViT-64-S/16 & Low-Rank Attention & 80.83 & 89.71 & 98.70 & 57.87 \\
	 Nystrom ViT-32-S/16 & Low-Rank Attention & 80.54 & 87.96 & 98.47 & 57.53 \\
	 Linformer-S/16 & Low-Rank Attention & 81.46 & 90.37 & 98.66 & 58.11 \\
	 XCiT-S-12/16 & Low-Rank Attention & 83.65 & 93.27 & 99.30 & 58.71 \\
	 SwinV2-Ti-W7/4 & Sparse Attention & 83.11 & 92.21 & 98.72 & 58.46 \\
	 Swin-Ti-W7/4 & Sparse Attention & 82.91 & 92.89 & 99.08 & 58.48 \\
	 Swin-S-W7/4 & Sparse Attention & 84.87 & 92.74 & 99.56 & 59.02 \\
	 Sinkhorn Cait-S-Bmax-32/16 & Sparse Attention & 81.27 & 92.79 & 98.48 & 57.64 \\
	 Routing ViT-S/16 & Sparse Attention & 73.92 & 84.06 & 97.37 & 55.44 \\
	 WaveViT-S & Sparse Attention & 83.91 & 92.69 & 99.33 & 57.72 \\
	 Informer-S/16 & Sparse Attention & 73.47 & 83.69 & 96.73 & 52.50 \\
	 Synthesizer FD-S/16 & Fixed Attention & 75.51 & 87.74 & 96.69 & 56.02 \\
	 Synthesizer FR-S/16 & Fixed Attention & 79.07 & 91.16 & 98.30 & 57.14 \\
	 PolySA-S/16 & Kernel Attention & 78.44 & 88.66 & 97.05 & 56.82 \\
	 SLAB-S/16 & Kernel Attention & 78.70 & 91.10 & 98.55 & 56.42 \\
	 Hydra-S/16 & Kernel Attention & 77.84 & 86.96 & 97.22 & 56.71 \\
	 EfficientFormerV2-S0 & Hybrid Attention & 71.53 & 89.51 & 95.55 & 53.43 \\
	 CvT-13 & Hybrid Attention & 82.35 & 90.72 & 98.70 & 58.17 \\
	 CoaT-Ti & Hybrid Attention & 78.42 & 90.45 &  & 55.41 \\
	 Efficient ViT-B-2 & Hybrid Attention & 81.52 & 88.29 & 97.36 & 56.85 \\
	 NextViT-S & Hybrid Attention & 83.92 & 92.90 & 98.89 & 58.28 \\
	 ResT-S & Hybrid Attention & 79.92 & 92.23 & 98.72 & 55.71 \\
	 Mixer-S/16 & Non-Attention Shuffling & 76.20 & 86.61 & 96.72 & 55.66 \\
	 FocalNet-S-Srf & Non-Attention Shuffling & 84.91 & 93.13 & 99.48 & 58.87 \\
	 FocalNet-Ti-Srf & Non-Attention Shuffling & 83.23 & 93.29 & 99.08 & 58.59 \\
	 SwiftFormer & Non-Attention Shuffling & 76.41 & 90.69 & 96.61 & 55.20 \\
	 FastViT-12 & Non-Attention Shuffling & 78.77 & 92.04 & 96.98 & 56.72 \\
	 Efficient Mod-S & Non-Attention Shuffling & 80.21 & 92.01 & 98.09 & 57.33 \\
	 FNet-S/16 & Fourier Attention & 73.12 & 87.87 & 95.52 & 54.64 \\
	 GFNet-S/16 & Fourier Attention & 81.33 & 91.92 & 98.80 & 57.97 \\
	 AFNO-S/16 & Fourier Attention & 78.55 & 90.30 &  & 56.09 \\
	 AFNO-Ti/16 & Fourier Attention & 68.87 &  &  &  \\
	 EViT-S/16 & Token Removal & 82.29 & 90.97 & 98.86 & 58.39 \\
	 Dynamic ViT-S/16 & Token Removal & 81.09 & 91.48 & 98.39 & 58.00 \\
	 EViT Fuse-S/16 & Token Merging & 82.20 & 90.53 & 98.75 &  \\
	 ToMe-S-R-8/16 & Token Merging & 82.11 & 91.58 & 98.61 & 58.20 \\
	 Token Learner-8-75-S/16 & Summary Tokens & 80.69 & 91.66 & 97.34 & 56.40 \\
	 Token Learner-8-50-S/16 & Summary Tokens & 77.35 & 90.29 & 94.56 & 54.37 \\
	 STViT-Swin-Ti-W-7/4 & Summary Tokens & 82.22 & 91.37 & 98.58 & 58.45 \\
	 CaiT-S-24 & Summary Tokens & 84.91 & 92.94 & 99.50 & 53.48 \\
	 Switch ViT-8-S/16 & More MLPs & 82.39 & 89.14 & 98.97 & 58.22 \\
	 Hi ViT-Ti/16 & More MLPs & 76.13 & 91.13 & 97.25 &  \\
	\cmidrule(r){1-2} \cmidrule(l){3-6}
	\cmidrule(r){1-2} \cmidrule(l){3-6}
	Correlation & with ImageNet-1k & 1.0 & 0.70 & 0.89 & 0.80 \\
	Spearman's $\rho$ & with ImageNet-1k & 1.0 & 0.71 & 0.93 & 0.83 \\
\bottomrule
\end{tabular}

\end{table*}

\begin{figure*}[ht!]
    \centering
    \includegraphics{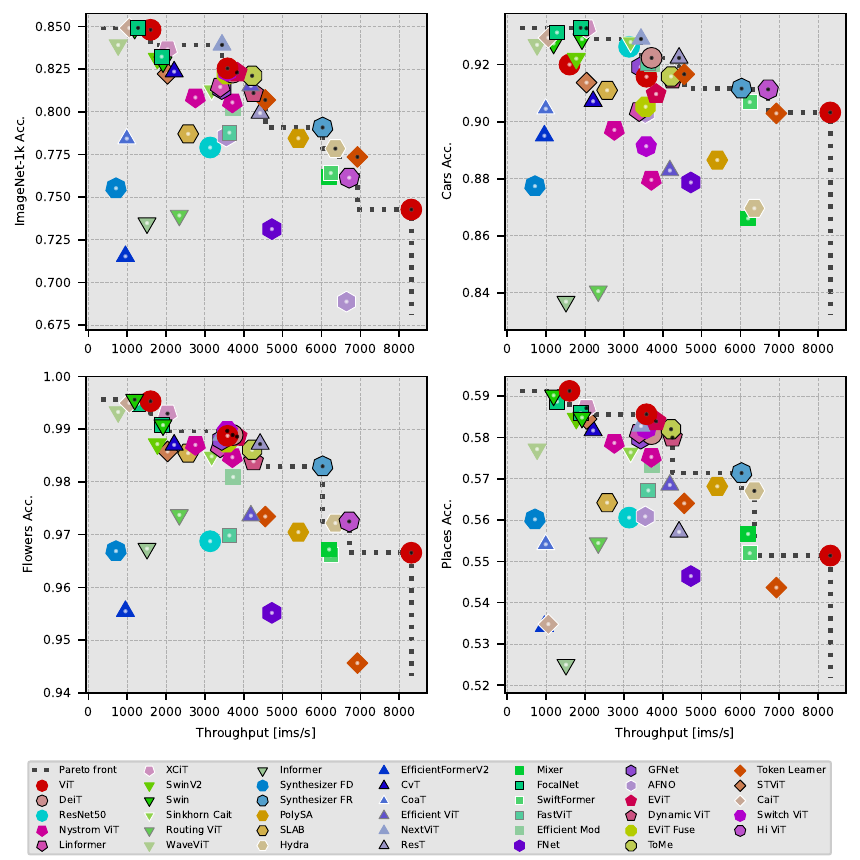}
    \caption{Pareto front of throughput and accuracy on four datasets for images of size $224 \times 224$ px. While the distribution of some models changes, the Pareto fronts are always very similar, independent of the dataset.}
    \label{fig:throughput_other_datasets}
\end{figure*}

We evaluate on image classification, as it is considered a challenging yet well-studied task to compare models.
However, we think that extending the benchmark to object detection and segmentation will give additional insights about efficiency in vision transformers.
We find this extension to be non-trivial, because of the added complexity of introducing a patch/token-level to pixel-level decoder and because some models remove/change tokens from the sequence, losing local information.
Therefore, evaluating on dense task falls out of the scope of our paper, and we consider it future work.

Additionally to ImageNet, we further finetune models on the Stanford Cars \cite{Krause2013}, Oxford Flowers 102 \cite{Nilsback2008}, and MIT Places365 \cite{Zhou2014} datasets using the hyperparameters in \Cref{tab:other_dataset_hyperparameters}.
Model accuracies for these down-stream tasks are presented in \Cref{tab:other_datasets}.
We observe a high correlation of accuracies between the down-stream and ImageNet performances, with even higher rank-correlations meaning that the order of models does \emph{not} significantly change from one dataset to another.
This observation is corroborated by \Cref{fig:throughput_other_datasets} where we plot the Pareto front of throughput and accuracy for the four different datasets.
Even though the distribution of some models changes, as they are finetuned on other tasks, the Pareto front stays largely the same, with differences probably stemming from the different number of training examples in the datasets.

\FloatBarrier
\section{Results: Details}

\subsection{Training Failures}
Despite our efforts to hyperparameter tune the models based on their original papers, we encounter training failures with several models using the new pipeline.
Specifically, we are unable to achieve convergence with the \emph{Performer} \cite{Choromanski2021}, \emph{Linear Transformer} \cite{Katharopoulos2020}, and \emph{HaloNet} \cite{Vaswani2021} models.

The \emph{Performer} and \emph{Linear Transformer} are found to be consistently unstable across all hyperparameter configurations, despite adjusting various training parameters such as learning rate, weight decay, and batch size.
In the case of \emph{HaloNet}, we find the implementation\footnote{based on the \texttt{halonet\_pytorch} package \url{https://github.com/lucidrains/halonet-pytorch}} to be prohibitively slow, and experiments indicate that the model does not exhibit meaningful learning even after multiple epochs of training on ImageNet-21k with different hyperparameters.
Specifically, after more than 10 epochs of training, the \emph{HaloNet} model is still performing no better than random choice, and due to the high computational cost of training, we do not pursue further hyperparameter tuning.
The \emph{Reformer} \cite{Kitaev2020} model needs too much memory to fit on 8 NVIDIA A100s (40GB), and we therefore did not train it.
Additionally, we only achieve convergence with \emph{A-ViT} \cite{Yin2022} using a very small learning rate, which results in suboptimal results within our fixed number of epochs.
We also don't reach satisfying results with \emph{GTP} \cite{Xu2023b} and therefore exclude it from our analysis.
Overall, we find it harder to get the kernel attention and some of the more complicated sequence reduction models to converge, than models of the other categories.

We note that these results may not be representative of the performance of these models, and further investigation is needed to understand the causes of the training failures.

\subsection{ImageNet Accuracy in Taxonomy Classes}
\label{apdx:acc-per-class}
\begin{figure*}[ht!]
    \centering
    \includegraphics{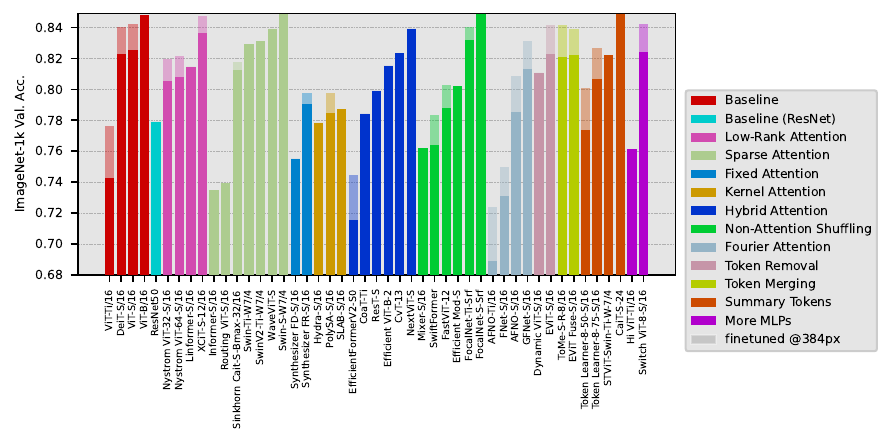}
    \caption{Accuracy of trained models trained, grouped by taxonomy classes and ordered by accuracy inside the classes. The base bar is the model trained at $224 \times 224$ pixels, while the lighter bar is the same model fine-tuned at $384 \times 384$ pixels, if training fits on 8 NVIDIA A100s and the accuracy is higher than the base model.}
    \label{fig:model_vs_acc}
\end{figure*}

\Cref{fig:model_vs_acc} shows the ImageNet accuracy achieved by the trained models. The models are categorized into distinct taxonomy classes. Notably, we observe that while models belonging to various taxonomy classes reach the highest accuracy levels, those falling under the \emph{fixed attention} and \emph{kernel attention} categories appear to demonstrate comparatively lower accuracy outcomes.
For \emph{fixed attention}, we hypothesize this may be due to the process of using the same attention matrix for every input image, while for \emph{kernel attention} it might be due to convergence difficulties or numerical problems during training.

\subsection{Speed on Different Hardware}
\label{apdx:speed-hardware}
\begin{table*}[ht!]
    \centering
    \small
    \caption{Correlation coefficient (left) and Spearman correlation (right) of measuring speed on different hardware -- NVIDIA A100 (A100), NVIDIA GeForce RTX 3090 (RTX 3090), and Intel i9-12900H (CPU).}
    \label{tab:speed_cor_table}
    \begin{tabular}{ccc}
	\begin{tabular}{lcc}
		\toprule
		Correlation & CPU & RTX 3090 \\
		\cmidrule(r){1-1} \cmidrule(l){2-3}
		A100 & 0.49 & 0.97 \\
		CPU &  & 0.75 \\
	\bottomrule
	\end{tabular}
& \quad &
	\begin{tabular}{lcc}
		\toprule
		Spearman $\rho$ & CPU & RTX 3090 \\
		\cmidrule(r){1-1} \cmidrule(l){2-3}
		A100 & 0.75 & 0.96 \\
		CPU &  & 0.80 \\
	\bottomrule
	\end{tabular}
\end{tabular}

\end{table*}
\begin{figure*}[ht!]
    \centering
    \includegraphics{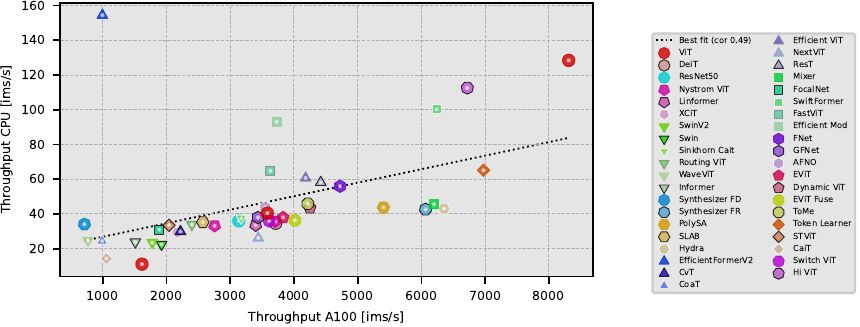}
    \caption{Throughput comparison on NVIDIA A100 and Intel i9-12900H (CPU) with line of best fit. Most models closely follow a linear relationship, with two very notable outliers. In particular \emph{EfficientFormerV2} \cite{Li2023d} is one of the slowest models on GPU, but one of the fastest ones on CPU.}
    \label{fig:throughput_gpu_vs_cpu}
\end{figure*}
\begin{figure*}[ht!]
    \centering
    \includegraphics{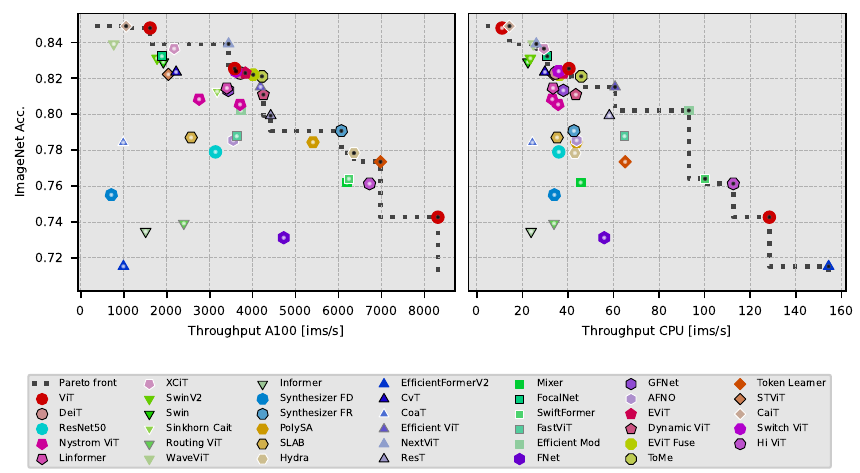}
    \caption{Pareto front of accuracy (ImageNet-1k) and throughput on NVIDIA A100 and Intel i9-12900H (CPU). We observe that the Pareto fronts are largely the same, with the one on CPU looking more squished because of the relatively larger speed of ViT-Ti, which we already observed in \Cref{fig:throughput_gpu_vs_cpu} to be an outlier.}
    \label{fig:pareto_throughput_gpu_vs_cpu}
\end{figure*}
We extend our analysis beyond large data center GPUs to CPUs and consumer GPUs.
\Cref{tab:speed_cor_table} shows a strong correlation between all the architectures with Spearman correlations exceeding 75\%.
While data center GPU and consumer GPU performance are 97\% correlated, GPU-CPU correlation is lower, with Spearman correlation between 75\% and 80\%.
When plotting GPU-throughput against CPU-throughput in \Cref{fig:throughput_gpu_vs_cpu}, we observe most models adhering to a linear relationship with some notable outliers: \emph{EfficientFormerV2}, \emph{HiViT}, \emph{SwiftFormer}, \emph{EfficientMod} and \emph{ViT-Ti}.
These models demonstrate significantly faster performance on CPU relative to GPU, while \emph{ViT-B} exhibits the opposite trend, being notably slower on CPU compared to GPU.
We hypothesize that these disparities stem from differences in memory access patterns between GPUs and CPUs.
While some very small models are faster on CPU than predicted by GPU performance, larger models get slower.
The presence of these outliers reduces the overall correlation to 49\%.
Without outliers, the correlation for the remaining models becomes 76\%.
A similar picture paints itself in \Cref{fig:pareto_throughput_gpu_vs_cpu}, where the overall Pareto front on CPU looks similar to the one on GPU, but a little squished because of the outstanding CPU-speed of \emph{ViT-Ti} and \emph{EfficientFormer}.
Again, we find \emph{ViT-S} and \emph{ViT-Ti} at the Pareto front, but not \emph{ViT-B}, due to the different scaling behavior on CPU.
Additionally, we find some of the outliers from before on the Pareto front.
Namely, \emph{EfficientFormerV2}, \emph{SwiftFormer}, \emph{HiViT}, and \emph{EfficientMod} are on the Pareto front on CPU, but not on GPU.
We again attribute this to different memory access properties on GPU and CPU, as \emph{Synthesizer} has a constant, learned attention matrix stored in memory.

\subsection{Model Scaling \& Image Scaling}
\begin{figure*}[ht!]
    \centering
    \includegraphics{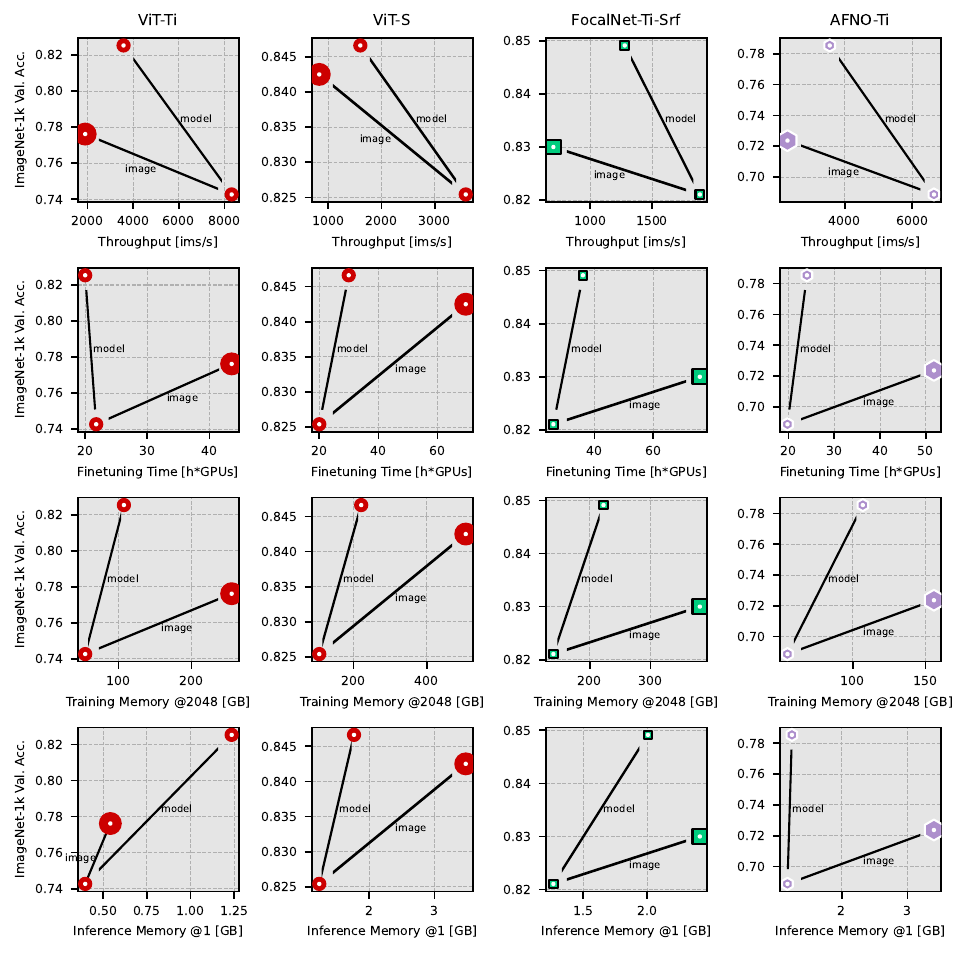}
    \caption{Effects of scaling the model size and the image resolution. Small markers use $224$px images, while large markers use $384$px images. Columns are different model combinations with titles marking the smaller model. For scaling the model, we use the next larger size. The rows are different dimensions of efficiency on the x-axes. We measure training memory at our default batch size of $2048$ and inference memory at the minimum possible batch size of $1$.}
    \label{fig:image_vs_model_scaling}
\end{figure*}

In \Cref{fig:image_vs_model_scaling} we show examples for the different effects of scaling the model size vs. scaling the image resolution.
For the three combinations of models \mbox{\emph{ViT-Ti} $\to$ \emph{ViT-S}}, \mbox{\emph{ViT-S} $\to$ \emph{ViT-B}}, \mbox{\emph{FocalNet-Ti} $\to$ \emph{FocalNet-S}}, and \mbox{\emph{AFNO-Ti} $\to$ \emph{AFNO-S}} (one per column) we plot the effects of scaling the image resolution at the smaller model size and scaling the model, but still using the smaller image size on the tradeoff of accuracy and one of the empirical efficiency metrics.
We observe that scaling the model size is almost always advantageous compared to scaling the image resolution.
The only exception being inference memory, where scaling the image resolution seems to work well for small models.
This is similar to the overall trend using larger images when optimizing for inference memory, that we found in the main paper.
Note, that in Figure 1 of the main paper, where we plot the Pareto front of all models and the Pareto front when only considering the models using high resolution images, it looks like both fronts are the same for high accuracy models.
This might be due to the fact that we did not include even larger models and therefore can not extend the Pareto fronts for even higher accuracy models.
In particular, the Pareto front in out plots goes flat at the top, where we do not have any more models to extend it.

\FloatBarrier
\subsection{Evolution of Models over Time}
\begin{figure*}[ht!]
    \centering
    \includegraphics{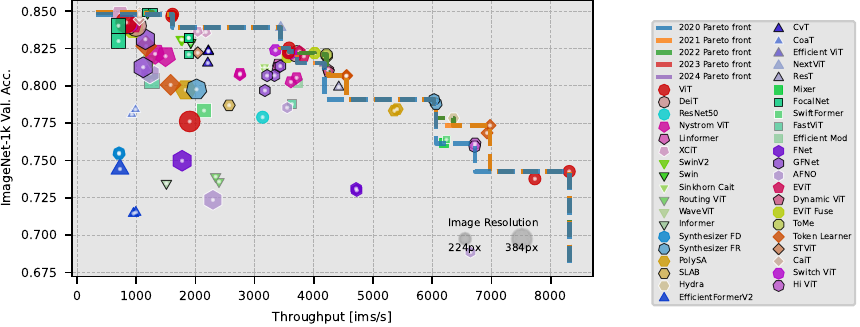}
    \caption{Pareto front of throughput and accuracy over time. Each fronts includes all models that were published up to that year. The improvements after 2020 are only marginal.}
    \label{fig:throughput_vs_acc_years_fronts}
\end{figure*}
\Cref{fig:throughput_vs_acc_years_fronts} shows the Pareto front of throughput and accuracy over time.
We observe that the Pareto front has not changed significantly since 2020, with only marginal improvements in throughput and accuracy.
In 2021, the TokenLearner was added and in 2022 Hydra, EViT, and ToMe were added to the Pareto front.

\FloatBarrier
\subsection{Pretraining Image Resolution}
\begin{figure*}[ht!]
    \centering
    \includegraphics{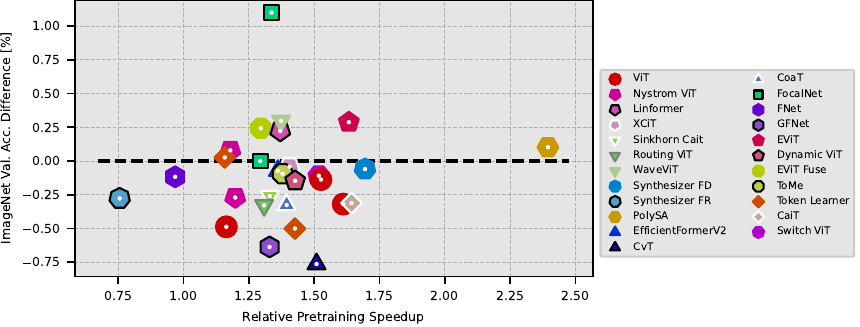}
    \caption{Impact of pretraining on ImageNet-21k at a lower resolution of $192$px instead of $224$px. Relative pretraining speedup is the ratio of pretraining time with $224$px images and $192$px images. All models are finetuned on ImageNet-1k at $224$px and we plot the difference of finetuned accuracy.}
    \label{fig:pretraining_image_resolution}
\end{figure*}
When pretraining at a lower resolution of $192$px images, we find an average decrease in the accuracy of $0.16\%$ after finetuning on ImageNet-1k at a resolution of $224$px for a $1.39$ times pretraining speedup on average. For most models, the speedup is between $1.1$ times and $1.6$ times. While accuracy increases for some models, for most it decreases with up to $0.8\%$ accuracy loss on ImageNet-1k.

\subsection{Accuracy per Parameter}
\label{apdx:acc-per-param}
\begin{figure*}[ht!]
    \centering
    \includegraphics{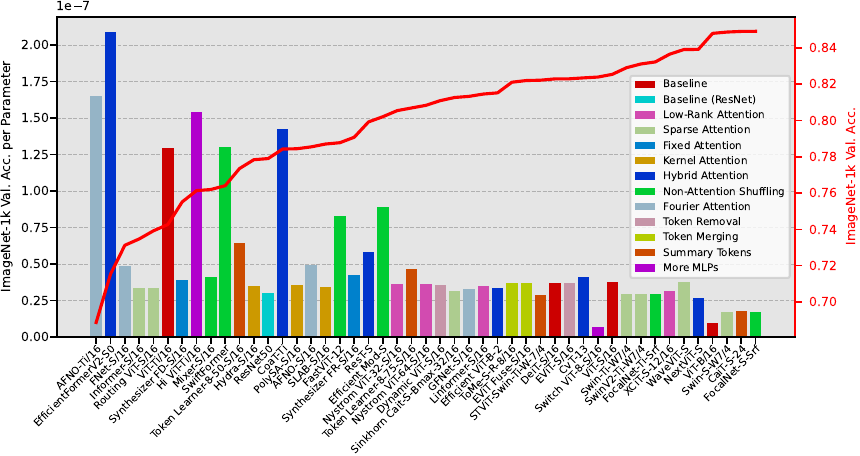}
    \caption{Accuracy (red line, right y-axis) and accuracy per parameter (bars, left y-axis) of different models at a resolution of $224 \times 224$ pixels. Models are ordered by accuracy.}
    \label{fig:model_vs_acc_per_param_and_acc}
\end{figure*}
\begin{figure*}[ht!]
    \centering
    \includegraphics{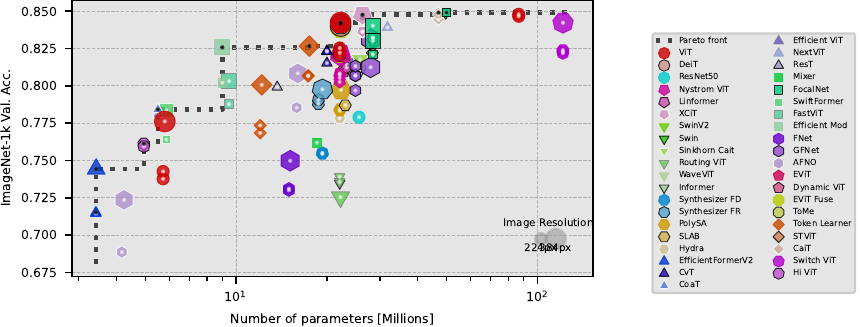}
    \caption{Pareto front of parameters and accuracy. We have a logarithmic x-axis in order to better show the differences between models. Since scaling to a larger image size is not represented in this metric, most Pareto optimal models use the larger image size.}
    \label{fig:parameters_vs_acc}
\end{figure*}
\Cref{fig:model_vs_acc_per_param_and_acc} shows the full list of models ordered by accuracy and the accuracy per parameter.
In the main paper, Synthesizer FR-S/16 to EViT-S/16 are collapsed into the \emph{others} bar, as their accuracy per parameter is very similar.

\Cref{fig:parameters_vs_acc} shows the Pareto front of plotting the number of parameters against accuracy.
In this metric, using a larger image size of $384$ does not show up and therefore is an advantageous strategy to increase accuracy.
Notably, the only Pareto optimal model that uses the smaller image size is CoaT-Ti.

\FloatBarrier
\subsection{Fine-tuning Time}
\label{apdx:finetuning-time}
\begin{figure*}[ht!]
    \centering
    \includegraphics{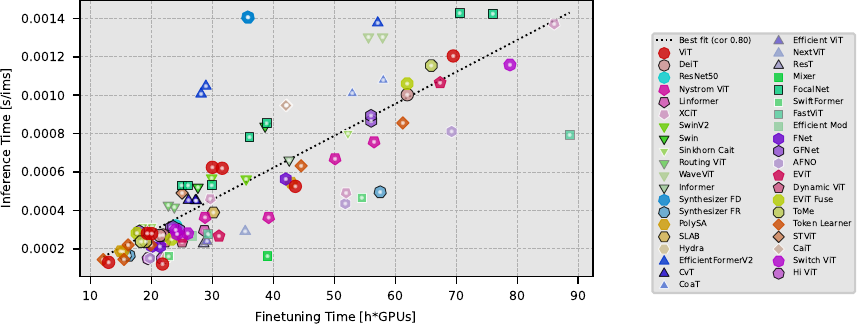}
    \caption{Fine-tuning time vs. inference time. This includes all runs at resolutions $224 \times 224$ and $384 \times 384$ pixels.}
    \label{fig:finetuning_time_vs_inference_time}
\end{figure*}
\begin{figure*}[ht!]
    \centering
    \includegraphics{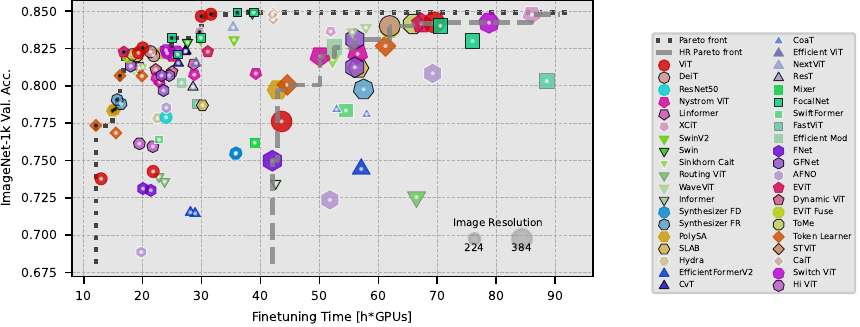}
    \caption{Pareto front of fine-tuning time vs. accuracy.}
    \label{fig:finetuning_time_vs_acc}
\end{figure*}

\Cref{fig:finetuning_time_vs_inference_time} visualizes the correlation of $0.8$ between fine-tuning time and inference time.
Except the few outliers at the top, this is especially pronounced for fast models and scatters more and more moving towards the slower ones.

Furthermore, \Cref{fig:finetuning_time_vs_acc} shows the Pareto front of finetuning time and accuracy, which is similar to the one on inference time and accuracy.
This similarity can be explained by the rather high correlation of finetuning time and inference time, we observed above.

\FloatBarrier
\subsection{Training Memory}
\begin{figure*}[ht!]
    \centering
    \includegraphics{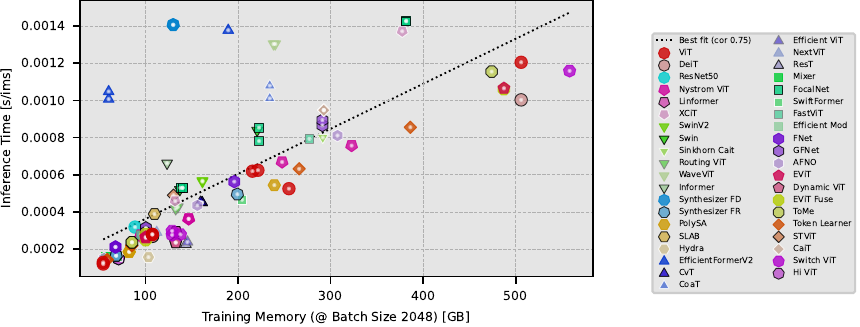}
    \caption{Training memory vs. inference time.}
    \label{fig:training_memory_vs_inference_time}
\end{figure*}

\begin{figure*}[ht!]
    \centering
    \includegraphics{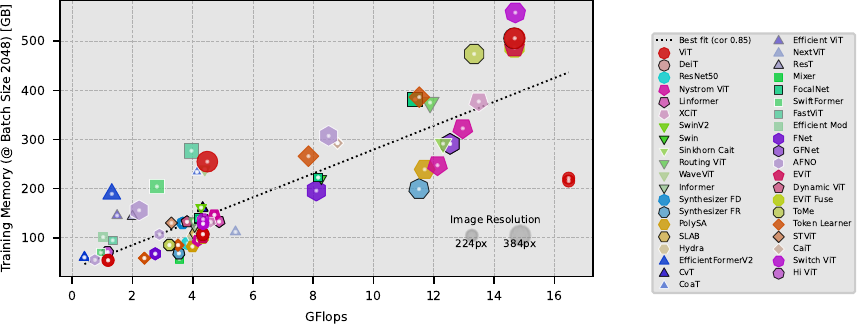}
    \caption{Flops vs. training memory.}
    \label{fig:flops_vs_training_memory_size_imsize}
\end{figure*}

\begin{figure*}[ht!]
    \centering
    \includegraphics{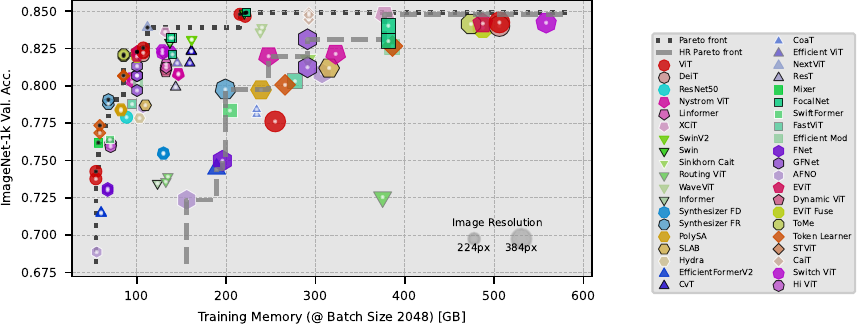}
    \caption{Full plot of training memory against accuracy. In the main paper, we only show the excerpt of $\leq 225$GB training memory.}
    \label{fig:training_memory_vs_acc_size_imsize}
\end{figure*}

\Cref{fig:training_memory_vs_inference_time} visualizes the correlation of $0.75$ between training memory and inference time.
This makes the Pareto front of training memory similar to the one of throughput.
Nonetheless, we observe some significant outliers, that need relatively little memory for training, but are slow.

\Cref{fig:flops_vs_training_memory_size_imsize} plots Flops against training memory of training runs at multiple resolutions.
A robust correlation coefficient of $0.85$ spans this diverse array of scenarios, enabling us to leverage the theoretical metric of throughput for an estimation of VRAM requirements during training, using the equation
\begin{align*}
    \text{VRAM [GB]} \approx 25.43 \cdot \text{GFlops} + 25.50.
\end{align*}
There are no significant outliers, similar to the level observed in \Cref{fig:training_memory_vs_inference_time}, to this relationship.

\Cref{fig:training_memory_vs_acc_size_imsize} shows the full plot of training memory against accuracy. In the main paper, we only include an excerpt.

\FloatBarrier
\subsection{Pareto Optimal Models}
\label{apdx:pareto-optimal-models}
\begin{table*}[ht!]
    \centering
    \caption{List of Pareto optimal models for speed and memory consumption and training and inference. Models are grouped by taxonomy class. Models annotated with $\uparrow$ are using $384$px images.}
    \label{tab:pareto_optimal_list}
    \small
    \begin{tabular}{lccccc}
        \toprule
                                & Number of     & \multicolumn{2}{c}{Speed} & \multicolumn{2}{c}{Memory}                                          \\
        \cmidrule(lr){3-4} \cmidrule(l){5-6}
                                & Architectures & Inference                 & Training                   & Inference         & Training           \\
        \midrule
        Baseline                & 3             & ViT                       & ViT                        &                   & ViT                \\
        \cmidrule{0-0}
        Low-Rank Attention      & 3             &                           & XCiT                       &                   &                    \\
        \cmidrule{0-0}
        Sparse Attention        & 6             &                           &                            &                   & Swin               \\
        \cmidrule{0-0}
        Fixed Attention         & 2             & Synthesizer-FR            & Synthesizer-FR             &                   & Synthesizer-FR     \\
        \cmidrule{0-0}
        Kernel Attention        & 3             & Hydra                     & PolySA                     &                   &                    \\
        \cmidrule{0-0}
        Hybrid Attention        & 6             & NextViT                   &                            & \makecell[c]{CoaT                      \\ EfficientViT \\ CvT \\ NextViT} & NextViT \\
        \cmidrule{0-0}
        Fourier Attention       & 3             &                           &                            &                   &                    \\
        \cmidrule{0-0}
        Non-Attention Shuffling & 5             & FocalNet                  & FocalNet                   & EfficientMod      & \makecell[c]{Mixer \\ FocalNet} \\
        \cmidrule{0-0}
        Token Removal           & 2             & \makecell[c]{EViT                                                                               \\ DynamicViT} & EViT & EViT@384 & EViT \\
        \cmidrule{0-0}
        Token Merging           & 2             & \makecell[c]{EViT Fuse                                                                          \\ ToMe} & & & ToMe \\
        \cmidrule{0-0}
        Summary Tokens          & 3             & TokenLearner              & TokenLearner               & CaiT              & TokenLearner       \\
        \cmidrule{0-0}
        More MLPs               & 2             &                           &                            &                                        \\
        \bottomrule
    \end{tabular}
\end{table*}

\Cref{tab:pareto_optimal_list} presents a comprehensive list of all Pareto optimal models for speed and memory consumption across both training and inference phases, grouped by taxonomy class.
We highlight select models from this list in the flowchart featured in the main paper.

\FloatBarrier

\end{document}